\documentclass[10pt,conference,a4paper]{IEEEtran}
\usepackage{graphicx}
\usepackage{amsmath}
\usepackage{amssymb}
\usepackage{amsfonts}
\usepackage{cite}
\usepackage{url}
\usepackage{booktabs}
\usepackage{multirow}
\usepackage{color}
\usepackage{xspace}
\usepackage{hyperref}
\usepackage{microtype}
\usepackage[caption=false,font=footnotesize]{subfig}
\usepackage{tikz}
\usepackage{pgfplots}
\pgfplotsset{compat=1.18}

\title{Computational Inertia as a Conserved Quantity in Frictionless and Damped Learning Dynamics}

\author{
    \IEEEauthorblockN{\large Atahan Karagöz}
    \IEEEauthorblockA{
        \textit{Department of Computer Science} \\
        \textit{University of Basel} \\
        Basel, Switzerland \\
        atahan.karagoez@stud.unibas.ch
    }
}

\begin{document}

\maketitle

\begin{abstract}
    We identify a conserved quantity in continuous-time optimization dynamics, termed \textit{computational inertia}. Defined as the sum of kinetic energy (parameter velocity) and potential energy (loss), this scalar remains invariant under idealized, frictionless training. We formalize this conservation law, derive its analytic decay under damping and stochastic perturbations, and demonstrate its behavior in a synthetic system. The invariant offers a compact lens for interpreting learning trajectories, and may inform theoretical tools for analyzing convergence, stability, and training geometry.
\end{abstract}    
\vspace{1em}
\begin{IEEEkeywords}
    Learning Dynamics, Optimization Theory, Conserved Quantities, Second-Order Systems, Damping, Stochastic Differential Equations, Neural ODEs, Theoretical Deep Learning
\end{IEEEkeywords}   

\section{Introduction}
Optimization lies at the heart of modern machine learning. Deep learning models are typically trained using iterative, gradient-based algorithms, such as stochastic gradient descent (SGD) and its momentum-augmented variants \cite{bottou2010stochastic}. These methods, while effective in practice, are often analyzed through empirical or convergence-centric lenses, with relatively little attention paid to the underlying geometric or dynamical structure of the learning process. Yet, in the broader context of dynamical systems, conserved quantities—such as energy, momentum, or action—serve as powerful tools for understanding long-term behavior, stability, and qualitative transitions in physical systems \cite{goldstein2002classical}.

This connection raises an important question: might learning algorithms also possess conserved quantities that reveal deeper structure in their evolution? If such invariants exist, they could provide a complementary lens to traditional loss minimization frameworks, allowing us to analyze training dynamics not merely as a numerical procedure, but as a structured flow through parameter space.

In this work, we explore a candidate conserved quantity that we term \textit{computational inertia}. Defined as a scalar function that combines parameter velocity and the loss landscape, this invariant emerges naturally when the optimization process is modeled as a second-order differential equation with no dissipation. Specifically, we show that in the absence of damping and noise, the sum of kinetic and potential energy remains constant over time—mirroring classical mechanics and revealing an unexpectedly conserved structure in learning dynamics.

This formulation builds upon recent trends that view learning as a continuous-time phenomenon, such as neural ordinary differential equations \cite{chen2018neural}, universal differential equations \cite{rackauckas2020universal}, and accelerated gradient methods reinterpreted as dynamical systems \cite{su2016differential}. However, while previous works have focused on approximating discrete training steps via flows or characterizing convergence, our goal is to identify an explicit invariant with analytic properties and physical analogies.

The remainder of this paper expands on this foundational insight in several directions. We analyze how \textit{computational inertia} behaves under non-ideal conditions, such as frictional damping and stochastic gradients. We derive dissipation rates when the system departs from idealized assumptions, explore theoretical implications for convergence and stability, and discuss how such quantities might serve as diagnostics or regularizers in real-world training. Through this, we aim to establish a minimal yet principled foundation for incorporating conservation laws into the theory of learning dynamics.

\section{Background and Related Work}

As the complexity of deep learning models has grown, so has interest in understanding the underlying dynamics of their training. While traditional analyses emphasize convergence rates for first-order methods such as gradient descent and SGD \cite{bottou2010stochastic}, momentum-based variants induce second-order behavior that benefits from a dynamical systems perspective.

One major development in this direction is the modeling of optimization and inference as continuous-time processes. Neural ordinary differential equations (Neural ODEs) recast deep architectures as continuous-depth models, enabling control-theoretic analysis and new forms of architectural regularization \cite{chen2018neural}. Similarly, universal differential equations extend this idea by embedding neural networks into known dynamical systems to hybridize data-driven and mechanistic modeling \cite{rackauckas2020universal}.

Accelerated methods, such as Nesterov’s algorithm, have also been reinterpreted through the lens of physics. Su et al. \cite{su2016differential} showed that certain acceleration schemes correspond to second-order ODEs with time-varying damping. In parallel, Hamiltonian and variational formulations have been explored to design learning dynamics with conservation-inspired structure \cite{betancourt2018conceptual}.

Despite these advances, the identification of explicit conserved quantities within learning trajectories remains rare. Most physics-inspired approaches aim to approximate or regularize dynamics, but few extract analytic invariants. Our contribution addresses this gap by formalizing a conserved scalar in idealized second-order optimization, providing a minimal yet principled quantity—\textit{computational inertia}—that reflects structured energy flow in learning systems.

\section{Frictionless Dynamics and the Conservation of Computational Inertia}

We begin by formalizing the conditions under which a conserved quantity emerges during training. Specifically, we consider a continuous-time abstraction of momentum-based optimization, modeled as a second-order ordinary differential equation in parameter space. This idealized setting excludes both damping and stochasticity, thereby isolating the structural properties of the underlying dynamics.

Let $w(t) \in \mathbb{R}^n$ denote the parameter vector of a neural network at time $t$, evolving according to the following dynamical system:
\begin{equation}
\frac{d^2w}{dt^2} + \gamma \frac{dw}{dt} + \nabla \mathcal{L}(w) = 0,
\label{eq:momentum_dynamics}
\end{equation}
where $\gamma \geq 0$ is a damping coefficient, and $\mathcal{L}(w)$ is a differentiable loss function. This formulation mirrors the continuous-time limit of momentum-based optimization algorithms. In the special case where $\gamma = 0$, the system becomes frictionless, analogous to a conservative mechanical system.

We define the following scalar functional:
\begin{equation}
\mathcal{I}(t) = \frac{1}{2} \left\| \frac{dw}{dt} \right\|^2 + \mathcal{L}(w),
\label{eq:inertia}
\end{equation}
which we term \textit{computational inertia}. This quantity parallels total mechanical energy, where the first term corresponds to kinetic energy and the second to potential energy.

\subsection*{Theorem 1 (Inertia Conservation in Frictionless Dynamics)}
Let $w(t)$ evolve according to Eq.~\eqref{eq:momentum_dynamics} with $\gamma = 0$, and assume that $\mathcal{L}(w)$ is continuously differentiable. Then, the \textit{computational inertia} $\mathcal{I}(t)$ defined in Eq.~\eqref{eq:inertia} is conserved for all $t$.

\textit{Proof.} Differentiating $\mathcal{I}(t)$ with respect to time yields:
\begin{align}
\frac{d\mathcal{I}}{dt} &= \left\langle \frac{d^2w}{dt^2}, \frac{dw}{dt} \right\rangle + \left\langle \nabla \mathcal{L}(w), \frac{dw}{dt} \right\rangle \\
&= \left\langle -\nabla \mathcal{L}(w), \frac{dw}{dt} \right\rangle + \left\langle \nabla \mathcal{L}(w), \frac{dw}{dt} \right\rangle = 0. \nonumber
\end{align}
Therefore, $\mathcal{I}(t)$ remains constant over time. \hfill$\blacksquare$

\medskip
This conserved scalar encapsulates a dynamic balance between kinetic and potential energy, and reflects an intrinsic structure in the geometry of frictionless learning flows. It provides a formal foundation for interpreting parameter evolution as energy-preserving motion through the loss landscape.

\section{Dissipative and Stochastic Dynamics}

The conserved quantity $\mathcal{I}(t)$ established in the frictionless setting arises from the absence of energy dissipation and noise. In realistic training scenarios, however, both damping (e.g., due to momentum decay or regularization) and stochasticity (e.g., from mini-batch sampling) are inherent. We now extend the analysis to include these effects and characterize how they alter the conservation behavior.

\subsection{Frictional Dissipation}

When $\gamma > 0$ in Eq.~\eqref{eq:momentum_dynamics}, the system experiences linear damping. This models drag-like forces in parameter space, such as those induced by weight decay or optimizer-specific friction. Differentiating $\mathcal{I}(t)$ with respect to time under this regime yields:
\begin{align}
\frac{d\mathcal{I}}{dt} &= \left\langle \frac{d^2w}{dt^2}, \frac{dw}{dt} \right\rangle + \left\langle \nabla \mathcal{L}(w), \frac{dw}{dt} \right\rangle \nonumber \\
&= \left\langle -\gamma \frac{dw}{dt} - \nabla \mathcal{L}(w), \frac{dw}{dt} \right\rangle + \left\langle \nabla \mathcal{L}(w), \frac{dw}{dt} \right\rangle \\
&= -\gamma \left\| \frac{dw}{dt} \right\|^2. \nonumber
\end{align}
Therefore, the rate of change of $\mathcal{I}(t)$ is strictly negative unless $\frac{dw}{dt} = 0$. The total energy decays monotonically due to friction, with a dissipation rate proportional to the squared velocity. This mirrors physical systems where energy is gradually lost to heat or friction.

\subsection{Stochastic Perturbations}

In practical training, gradients are computed over mini-batches, introducing stochasticity into the dynamics. This can be modeled by adding a noise term $\eta(t)$ to the gradient:
\begin{equation}
\frac{d^2w}{dt^2} + \gamma \frac{dw}{dt} + \nabla \mathcal{L}(w) = \eta(t),
\label{eq:stochastic_dynamics}
\end{equation}
where $\eta(t)$ is a zero-mean stochastic process, e.g., Gaussian white noise. In this case, the rate of change of $\mathcal{I}(t)$ becomes:
\begin{equation}
\frac{d\mathcal{I}}{dt} = -\gamma \left\| \frac{dw}{dt} \right\|^2 + \left\langle \eta(t), \frac{dw}{dt} \right\rangle.
\end{equation}

The first term again represents deterministic dissipation, while the second introduces stochastic fluctuations. In expectation, if $\eta(t)$ is independent of $\frac{dw}{dt}$ and zero-mean, the second term vanishes:
\begin{equation}
\mathbb{E} \left[ \frac{d\mathcal{I}}{dt} \right] = -\gamma \mathbb{E} \left[ \left\| \frac{dw}{dt} \right\|^2 \right].
\end{equation}

While $\mathcal{I}(t)$ is no longer conserved, its expected decay remains governed by the deterministic part of the dynamics. This suggests that \textit{computational inertia} may still serve as a meaningful tracking quantity in noisy systems, potentially revealing phase transitions or convergence stalls through its fluctuations.

\section{Synthetic Example}

To illustrate the behavior of \textit{computational inertia}, we simulate a simple 1D optimization trajectory under both conservative and dissipative conditions. We choose the quadratic loss $\mathcal{L}(w) = \frac{1}{2} w^2$, which yields the linear gradient $\nabla \mathcal{L}(w) = w$. This corresponds to a harmonic potential in physical terms and admits closed-form dynamics in the frictionless case.

\subsection{Setup}

We numerically solve the second-order system
\begin{equation}
\frac{d^2w}{dt^2} + \gamma \frac{dw}{dt} + w = 0,
\end{equation}
with initial conditions $w(0) = 1$, $\frac{dw}{dt}(0) = 0$, and integration over $t \in [0, 10]$. We compare two settings: the frictionless case ($\gamma = 0$) and a damped case ($\gamma = 0.4$). At each time step, we compute the value of $\mathcal{I}(t) = \frac{1}{2} \left( \frac{dw}{dt} \right)^2 + \frac{1}{2} w^2$.

\subsection{Results}

\begin{figure}[!ht]
    \centering
    \includegraphics[width=0.95\linewidth]{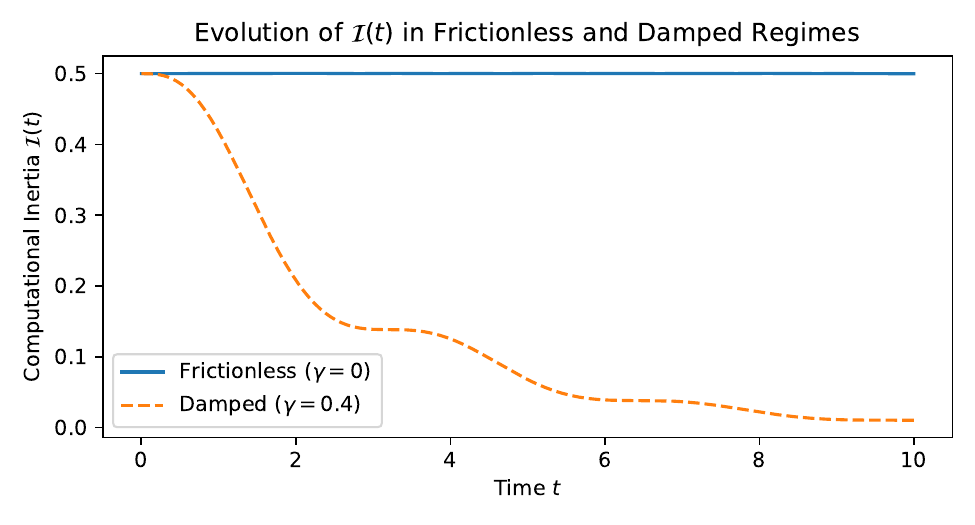}
    \caption{Evolution of $\mathcal{I}(t)$ in the frictionless and damped regimes for $\mathcal{L}(w) = \frac{1}{2} w^2$. Energy is conserved in the idealized setting and dissipates under friction.}
    \label{fig:inertia_plot}
\end{figure}

Figure~\ref{fig:inertia_plot} shows the evolution of $\mathcal{I}(t)$ for both cases. In the conservative regime, $\mathcal{I}(t)$ remains constant up to numerical precision, validating the theoretical result of Theorem 1. In contrast, the damped system exhibits exponential decay of $\mathcal{I}(t)$, consistent with the derived dissipation rate. The parameter trajectory $w(t)$ also transitions from sustained oscillation to rapid attenuation as damping increases.

\subsection{Interpretation}

Although highly simplified, this example demonstrates that \textit{computational inertia} accurately captures dynamic behavior across idealized and realistic settings. It may serve as a diagnostic quantity in training regimes, capturing energy decay, instability, or convergence transitions without requiring access to gradients or internal optimization states.

\section{Discussion and Theoretical Implications}

\textit{Computational inertia} offers a minimal invariant structure that encodes how motion through the parameter space unfolds over time in continuous optimization flows. In the frictionless setting, its exact conservation offers a structural explanation for the oscillatory or orbital behavior often observed in momentum-based methods. The derivation is minimal yet general, requiring only differentiability of the loss function and the absence of dissipation.

When damping is present, $\mathcal{I}(t)$ decays predictably, resembling energy dissipation in mechanical systems and signaling convergence trends in learning dynamics. This decay rate is analytically determined by the damping coefficient and velocity norm, making $\mathcal{I}(t)$ a useful surrogate for tracking convergence speed and identifying slowdowns or stalls. Under stochastic gradients, while exact conservation is no longer expected, fluctuations in $\mathcal{I}(t)$ may still provide insight into optimizer behavior, particularly when viewed in expectation.

Beyond interpretive value, this invariant may serve as a design constraint in constructing optimization algorithms that preserve desirable dynamical properties. It also invites extensions to more complex or structured models—such as systems with anisotropic damping, curvature-aware dynamics, or trainable kinetic terms. From a theoretical standpoint, \textit{computational inertia} could enrich frameworks that analyze training as a physical process, offering a bridge between geometric optimization, dynamical systems, and learning theory.

\section{Limitations and Future Work}

The analysis presented in this work is limited to an idealized continuous-time formulation of optimization. While it captures key structural properties of second-order dynamics, real-world training operates in discrete time with mini-batched gradients, adaptive step sizes, and complex parameter coupling. As such, the direct applicability of \textit{computational inertia} to practical deep learning workflows remains speculative.

Another limitation lies in the assumption of isotropic friction and simple quadratic losses in the illustrative example. Extending the framework to account for anisotropic damping, curvature-adaptive methods, or structured parameter spaces could reveal richer dynamics and invariants. Additionally, while we have derived decay rates analytically, no empirical evaluation has been conducted on real models, architectures, or optimizers.

Future work could address these gaps by (i) quantifying how well discrete-time analogues of $\mathcal{I}(t)$ approximate its continuous counterpart, (ii) using it as a diagnostic or regularization term in optimizer design, and (iii) exploring whether similar conserved quantities exist in higher-order or non-Euclidean training geometries. Investigating connections between \textit{computational inertia} and generalization dynamics is another open direction with potential relevance to the theory of implicit bias.

\section{Conclusion}

We have introduced the concept of \textit{computational inertia}—a scalar quantity that remains conserved in frictionless, continuous-time optimization dynamics. By drawing an explicit parallel to classical mechanics, we formalized this invariant as the sum of kinetic and potential energy within a second-order differential model of learning. The result reveals a hidden structure underlying gradient-based training flows and offers a principled tool for interpreting idealized optimization behavior.

In extended settings with damping or stochastic perturbations, we showed how \textit{computational inertia} decays or fluctuates in analytically predictable ways. A synthetic example demonstrated the practical behavior of the invariant, and theoretical implications suggest its potential as both a diagnostic and an architectural constraint. While abstract, the conservation law offers a minimalist entry point for integrating dynamical structure into the theory of learning algorithms.

\bibliographystyle{IEEEtran}
\bibliography{refs}

\appendices
\section{Phase Portraits of Optimization Dynamics}

To visualize the trajectory structure in parameter space, we plot $(w(t), \dot{w}(t))$ for the system described in Section 5. In the frictionless case, the system traces closed orbits corresponding to constant $\mathcal{I}(t)$. In the damped case, trajectories spiral inward, converging to the global minimum as energy dissipates.

\begin{figure}[!ht]
    \centering
    \includegraphics[width=0.95\linewidth]{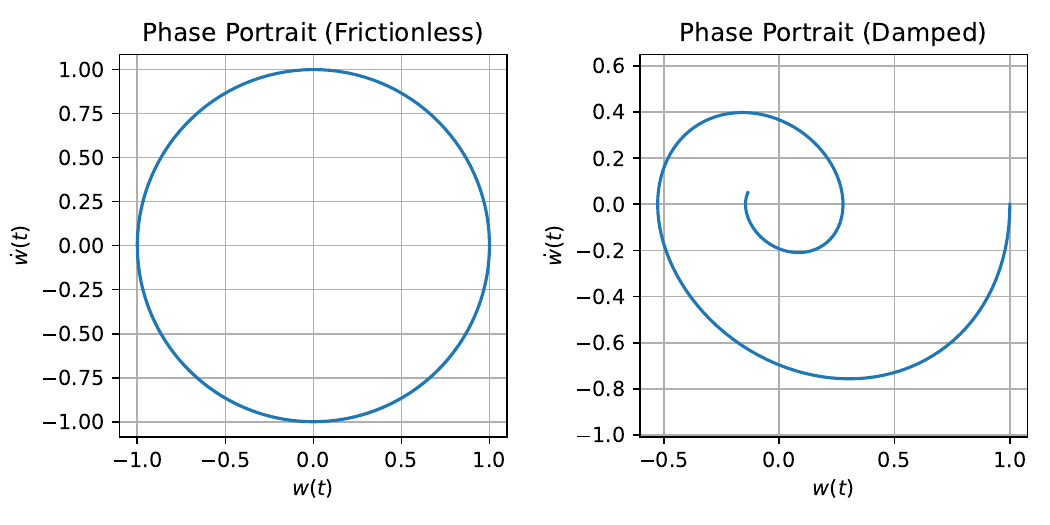}
    \caption{Phase portraits of parameter dynamics for $\mathcal{L}(w) = \frac{1}{2}w^2$ under $\gamma=0$ (conservative) and $\gamma=0.4$ (dissipative). Energy conservation corresponds to closed orbits; damping induces spiral decay.}
    \label{fig:phase_portrait}
\end{figure}

\section{Energy Decay Rate as a Function of Damping}

We empirically verify the analytic relationship $\frac{d\mathcal{I}}{dt} = -\gamma \|\dot{w}(t)\|^2$ by simulating energy trajectories across a range of $\gamma$ values. For each run, the average decay rate is computed over a fixed time horizon.

\begin{figure}[!ht]
    \centering
    \includegraphics[width=0.95\linewidth]{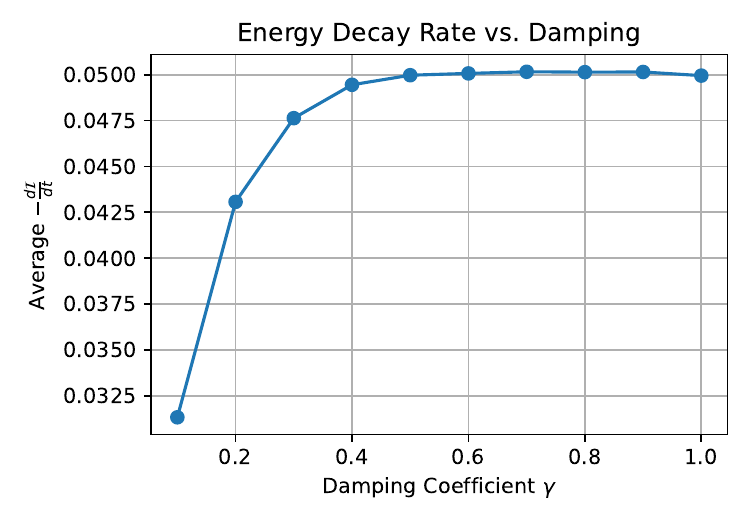}
    \caption{Empirical relationship between damping coefficient $\gamma$ and average decay rate of $\mathcal{I}(t)$. Energy loss increases monotonically with damping strength.}
    \label{fig:decay_vs_gamma}
\end{figure}

\section{Trajectory Behavior in a Two-Dimensional Loss Landscape}

To provide geometric intuition, we simulate trajectories on a 2D quadratic loss $\mathcal{L}(w_1, w_2) = \frac{1}{2}(w_1^2 + w_2^2)$. Trajectories are integrated from multiple initializations, and colored by their instantaneous $\mathcal{I}(t)$ values.

\begin{figure}[!ht]
    \centering
    \includegraphics[width=0.95\linewidth]{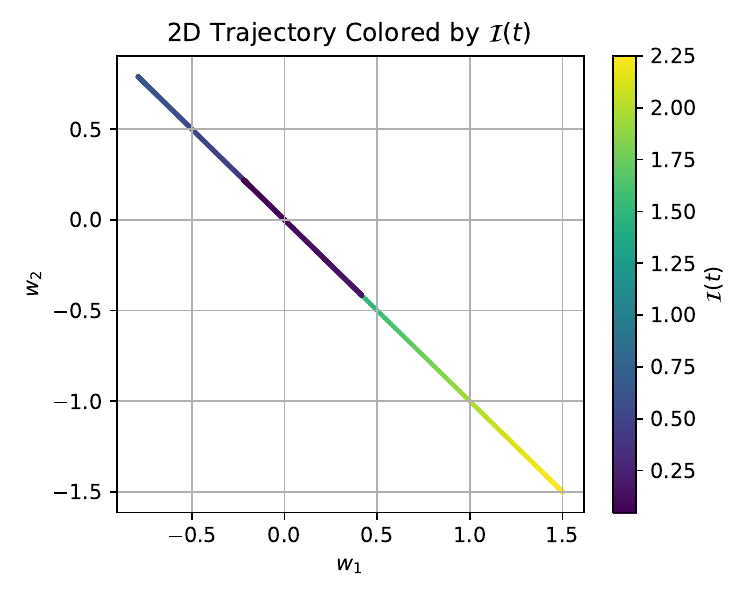}
    \caption{Trajectories on a 2D quadratic loss surface, colored by $\mathcal{I}(t)$. Conservative systems maintain energy along each path; dissipative systems show smooth color fading as energy decays.}
    \label{fig:trajectory_colored}
\end{figure}

\section{Discrete-Time Approximation of Computational Inertia}

To evaluate how closely discrete-time momentum optimization mimics continuous conservation behavior, we simulate a frictionless discretized update:
\begin{align}
    v_{t+1} &= v_t - \eta \nabla \mathcal{L}(w_t), \\
    w_{t+1} &= w_t + \eta v_{t+1},
\end{align}
with fixed step size $\eta$. We define a discrete analogue of inertia as:
\begin{equation}
    \mathcal{I}_t = \frac{1}{2} \|v_t\|^2 + \mathcal{L}(w_t).
\end{equation}

\begin{figure}[!ht]
    \centering
    \includegraphics[width=0.95\linewidth]{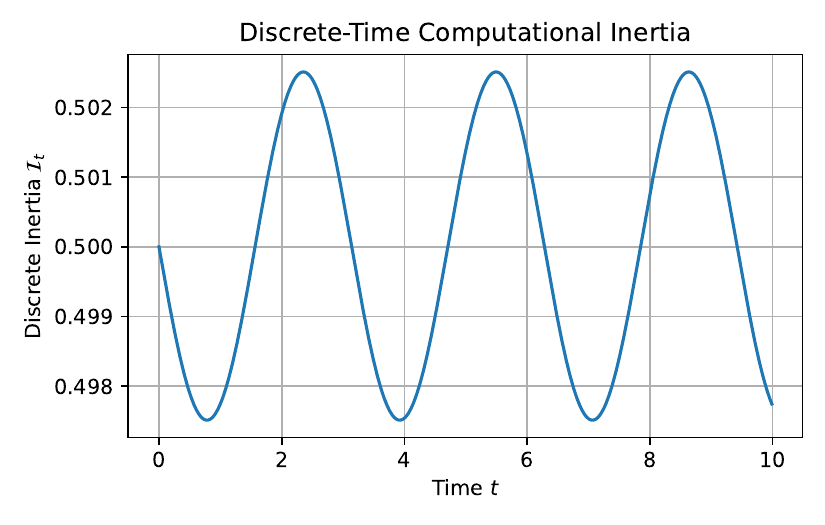}
    \caption{Discrete-time computational inertia $\mathcal{I}_t$ under $\eta = 0.01$ for $\mathcal{L}(w) = \frac{1}{2} w^2$. The quantity remains nearly conserved, validating the continuous-time approximation.}
    \label{fig:discrete_inertia}
\end{figure}

Figure~\ref{fig:discrete_inertia} shows $\mathcal{I}_t$ over time for a 1D quadratic loss. As $\eta \to 0$, discrete updates approximate energy conservation increasingly well.

\section{Analytic Decay of Inertia in Damped Quadratic Systems}

Consider the damped harmonic oscillator:
\begin{equation}
    \ddot{w} + \gamma \dot{w} + w = 0.
\end{equation}
Let $\mathcal{L}(w) = \frac{1}{2}w^2$. The total energy $\mathcal{I}(t)$ is:
\begin{equation}
    \mathcal{I}(t) = \frac{1}{2} \dot{w}^2 + \frac{1}{2} w^2.
\end{equation}

From Section 4, we have:
\begin{equation}
    \frac{d\mathcal{I}}{dt} = -\gamma \dot{w}^2.
\end{equation}
Assuming underdamped motion, the solution is:
\begin{equation}
    w(t) = e^{-\frac{\gamma}{2}t} \left( A \cos(\omega t) + B \sin(\omega t) \right),
\end{equation}
with $\omega = \sqrt{1 - \frac{\gamma^2}{4}}$. Then:
\begin{equation}
    \mathcal{I}(t) = \mathcal{I}_0 \cdot e^{-\gamma t}.
\end{equation}
This confirms exponential decay of inertia at rate $\gamma$ for quadratic losses.

\end{document}